\documentclass{article}
\usepackage{spconf,amsmath,amssymb,graphicx}

\usepackage{subcaption,booktabs}
\usepackage{array}[=2024/01/01]
\usepackage{multirow,arydshln}
\usepackage{algorithm,algpseudocode}
\usepackage{xcolor}
\usepackage{microtype}
\usepackage{balance}

\usepackage{pifont}%

\newcommand{\red}[1]{\textcolor{red}{#1}}
\newcommand{\blue}[1]{\textcolor{blue}{#1}}
\newcommand{\rev}[1]{\textcolor{purple}{#1}}
\newcommand{\cut}[1]{\textcolor{green}{#1}}
\renewcommand{\rev}[1]{#1}
\renewcommand{\cut}[1]{#1}

\newcommand{\ie}[1]{\textit{i.e.}#1}
\newcommand{\eg}[1]{\textit{e.g.}#1}

\let\OLDthebibliography\thebibliography
\renewcommand\thebibliography[1]{
  \OLDthebibliography{#1}
  \setlength{\parskip}{0pt}
  \setlength{\itemsep}{0pt plus 1ex}
}

\title{Continual Road-Scene Semantic Segmentation via Feature-Aligned Symmetric Multi-Modal Network}

\name{Francesco Barbato, Elena Camuffo, Simone Milani, Pietro Zanuttigh\thanks{\hspace{-1em}\{\makebox{francesco.barbato,elena.camuffo,simone.milani,zanuttigh}\}@dei.unipd.it. \rev{This work was partially supported by the European Union under the Italian National Recovery and Resilience Plan (NRRP) of NextGenerationEU, partnership on ``Telecommunications of the Future'' (PE00000001- program ``RESTART'').}}}

\address{University of Padova \\
Department of Information Engineering \\
Via Gradenigo 6/b, Padova, Italy}

\begin{document}
\maketitle

\begin{abstract}
State-of-the-art multimodal semantic segmentation strategies combining LiDAR and color data are usually designed on top of asymmetric information-sharing schemes and assume that both modalities are always available.
This strong assumption may not hold in real-world scenarios, where sensors are prone to failure or can face adverse conditions that make the acquired information unreliable. This problem is exacerbated when continual learning scenarios are considered since they have stringent data reliability constraints.
In this work, we re-frame the task of multimodal semantic segmentation by enforcing a tightly coupled feature representation and a symmetric information-sharing scheme, which allows our approach to work even when one of the input modalities is missing. We also introduce an ad-hoc class-incremental continual learning scheme, proving our approach's effectiveness and reliability even in safety-critical settings, such as autonomous driving. We evaluate our approach on the SemanticKITTI dataset, achieving impressive performances. 
\end{abstract}

\begin{keywords}
Continual Learning, Multimodal Learning, Semantic Segmentation, Scene Understanding
\end{keywords}

\begin{figure}[t]
    \centering
    \includegraphics[width=0.88\linewidth]{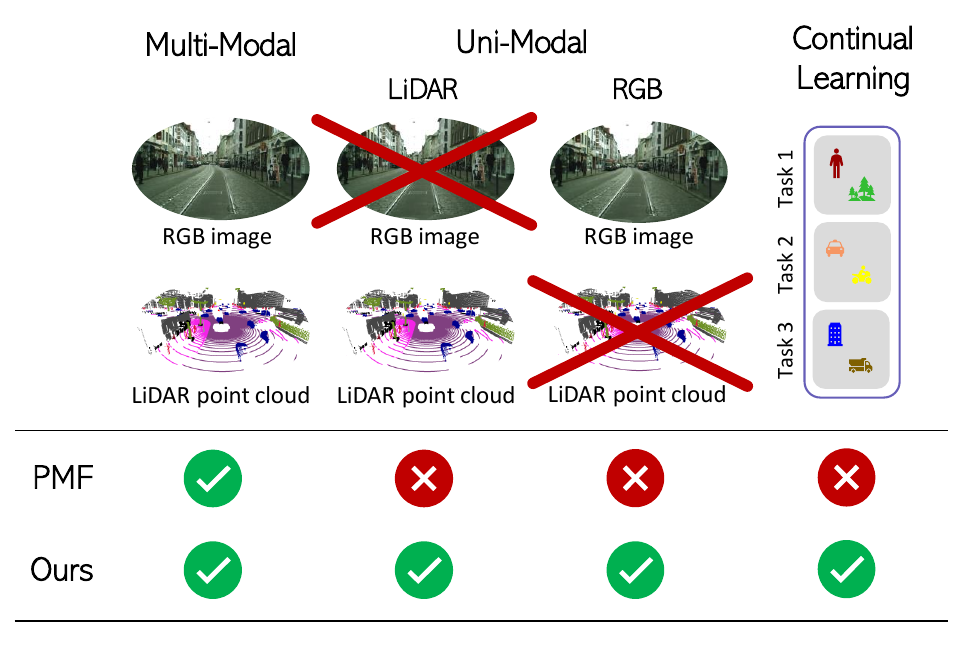}
    \caption{\small We introduce a new symmetric multimodal architecture that, differently from competitors \cite{zhuang2021perceptionaware}, works reliably even when one modality is missing. Our regularization strategies also allow for a natural extension to continual learning scenarios, where we relax the restriction of needing both modalities at the same time.}\label{fig:graphabs}
\end{figure}

\section{Introduction}
The environment perception in %
autonomous driving systems has recently improved thanks to the availability of multiple heterogeneous sensors including stereo cameras, depth sensors, LiDAR, and RADAR devices. Each comes with different accuracy, %
adverse weather robustness, and power consumption. %
\rev{Generally}, single-sensor systems face difficulties in adapting to different working conditions, especially when %
tackling complex %
problems such as semantic segmentation. %
To solve the issue, multimodal architectures have recently started appearing \cite{technologies10040090}, fostered by the availability of suitable datasets \cite{geiger2012cvpr,behley2019iccv,testolina2023selma,caesar2020nuscenes,huang2018apolloscape}.
Despite the improvements, most solutions make assumptions that limit applicability in a realistic scenario.
A common one presumes that a sensing device is dominant over the others, which are simply required 
to aid in uncertain situations.
Furthermore, most setups assume consistent noise and reliability between sensors, lack of environmental heterogeneity, and constant label distribution during training. 
All of these are unreasonable in the real world, where an agent may drive in new places or encounter new classes.
Here we present a color and LiDAR multimodal semantic segmentation architecture able to combine the two sensors, focusing on one modality over the other, depending on their availability. 
Following other state-of-the-art approaches for LiDAR semantic segmentation \cite{zhuang2021perceptionaware},  point clouds are aligned to the input images, and the two are fed to an architecture with a tightly coupled cross-modal feature representation, which - differently from other approaches - allows it to work even when one of the modalities is missing or corrupted. Furthermore, it can handle the label distribution shifts introduced by continual learning.
Remarkably, the proposed strategy matches state-of-the-art competitors while allowing the adoption of continual learning strategies and single-modality inputs. 
The major contributions of this work are the following: 
(i) We propose a symmetric multimodal semantic segmentation architecture where both or just one of the modalities can be used as input. To the best of our knowledge, this is the first fully balanced architecture that works indifferently either on LiDAR and RGB data or both. 
(ii) %
We investigate the reliability in a continual learning setting by introducing three knowledge distillation strategies (intra-modality, inter-modality, and cross-modality) and a multimodal inpainting approach.
(iii) %
We validate the architecture in the SemanticKITTI benchmark \cite{behley2019iccv} matching state-of-the-art performance of LiDAR-focused approaches while providing much greater flexibility.

\section{Related Works}\label{sec:related}
\textbf{Multimodal Semantic Segmentation} %
exploits multimodal data to obtain a dense (pixel-level) segmentation of an input scene \cite{melotti2020,li2022deepfusion,ASVADI201820}.
Early attempts %
combined RGB data and other modalities into multi-channel representations that were then fed into classical semantic segmentation %
models based on the encoder–decoder framework \cite{technologies10040090, barbato2023depthformer}.

Combining RGB images with LiDAR sensor data %
is a challenging problem due to the sparse and irregular structure of LiDAR point clouds.
A possible strategy is %
merging multiple LiDAR sensors \cite{Pfeuffer2019RobustSS} to obtain a denser cloud, although most existing approaches project 3D points %
over the color frame to %
fuse cross-modality features.
FuseSEG-LiDAR \cite{krispel2019fuseseg} adopts multi-layer concatenation of RGB-related features %
and spherically projects LiDAR points to generate an RGBD representation.
\rev{Perception-aware Multi-sensor Fusion \cite{zhuang2021perceptionaware} (PMF)} follows from this approach but changes the projection to perspective and the architecture to use residual-based fusion modules toward the LiDAR branch. 
Other approaches make the network learn both 2D textural appearance and 3D structural features in a unified framework \cite{jaritz2020xmuda}. Cheng \textit{et al.} \cite{unified2019} proposed a method in which they back-project 2D image features into 3D coordinates, while \cite{jaritz2019} aggregates 2D multi-view image features into 3D point clouds, and then uses a point-based network to fuse the features in 3D  space to predict \cut{3D semantic labels}.

\textbf{Continual Learning} literature in Semantic Segmentation is mostly concerned with RGB %
images and %
the application to multimodal data is still in an early stage \cite{Camuffo_2023_CVPR}.
Major attention has been devoted to class incremental semantic segmentation to learn new categories in subsequent steps \cite{cermelli2020modeling,klingner2020class,douillard2021plop,cermelli2022incremental,s22041357,Michieli_2021_SDR}. The problem is generally tackled with regularization approaches such as parameters freezing and knowledge distillation \cite{michieli2019}, which can be coupled with a class importance weighting scheme to emphasize gradients on difficult classes \cite{klingner2020class}. Also, latent space regularization is often used to improve class-conditional feature separation \cite{Michieli_2021_SDR}. 

\section{Problem Formulation}\label{sec:problem}

\begin{figure*}
    \centering
    \includegraphics[trim=0cm 0cm 0cm 0cm,clip,width=0.8\textwidth]{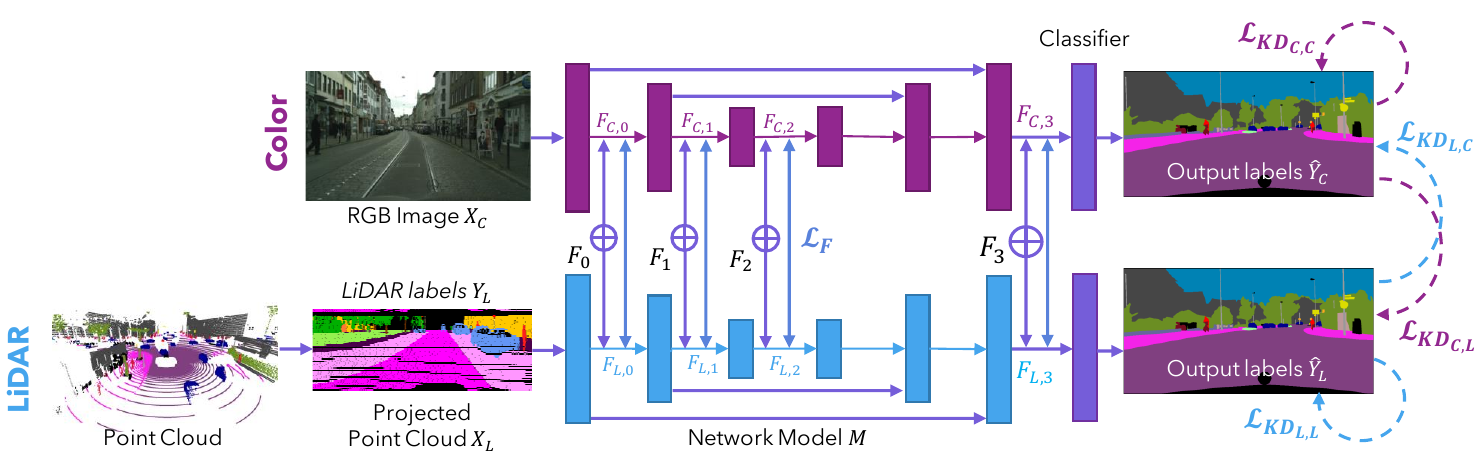}
    \caption{\small Overall architecture of our method %
    taking in input an RGB and LiDAR sample in different branches, aligning features at multiple levels, then combining the output from the two network branches to provide a single joint segmentation map. The   \textit{catastrophic forgetting} in continual learning is tackled via %
    knowledge distillation losses that help the modalities jointly by sharing information from the other. %
    }\label{fig:pipeline}
    \vspace{-1.1em}
\end{figure*}

Semantic segmentation is the task of assigning to each input sample $\mathbf{X}$ (\eg, images or 3D data) a dense labeling that approximates the ground truth $\mathbf{Y}$ in such a way that $\text{argmax}\{\hat{\mathbf{Y}}\} \sim \mathbf{Y} \in \mathcal{C}^{H \times W}$, where $\mathcal{C}$ is the set of known classes, while $H$ and $W$ %
are the image dimensions. The task is usually performed using an encoder-decoder-based deep neural network $M = D \circ E$.

In a multimodal setup, the input is given as an RGB-LiDAR pair %
, \ie, $(\mathbf{X}_C, \mathbf{X}_L) \in \mathcal{X}_C \times \mathcal{X}_L \subset \mathbb{R}^{H \times W \times 3} \times \mathbb{R}^{H \times W \times 5}$.
Note that, in our scenario, we allow also the possibility for one of the modality samples to be missing. Consequently, our training data can be provided in three different forms, depending on which modalities are present and which are missing: $(\varnothing,\mathbf{X}_L)$, $(\mathbf{X}_C, \varnothing)$, and $(\mathbf{X}_C, \mathbf{X}_L)$. 
\\
The model $M$ is defined as a 2-branch architecture (color branch $M_C$ + LiDAR branch $M_L$), which allows information sharing at the feature level in a multi-scale manner ($\mathbf{F}_{C,i}, \mathbf{F}_{L,i}$). Indeed, each of the two branches produces \rev{3 intermediate feature representations $(\mathbf{F}_{\cdot,0}, \mathbf{F}_{\cdot,1}, \mathbf{F}_{\cdot,2})$ and a final embedding ($\mathbf{F}_{\cdot,3})$ of the input: $F_C(\mathbf{X}_{C}) = (\mathbf{F}_{C,0}, \mathbf{F}_{C,1}, \mathbf{F}_{C,2}, \mathbf{F}_{C,3})$ for the color branch and $F_L(\mathbf{X}_{L}) = (\mathbf{F}_{L,0}, \mathbf{F}_{L,1}, \mathbf{F}_{L,2}, \mathbf{F}_{L,3})$ for the LiDAR branch}. 
Such features are fed to two separate decoders (one per branch). The color branch provides predictions $\mathbf{\hat{Y}}_{C} = D_C(\mathbf{F}_{C,0}, \mathbf{F}_{C,1},\allowbreak \mathbf{F}_{C,2}, \mathbf{F}_{C,3})$, the LiDAR branch provides predictions $\mathbf{\hat{Y}}_{L} = D_L(\mathbf{F}_{L,0}, \mathbf{F}_{L,1}, \mathbf{F}_{L,2}, \mathbf{F}_{L,3})$. 

The dataset used for %
validation provides %
ground truth %
for the LiDAR data only; hence we evaluate the output of the model using %
this representation \cite{zhuang2021perceptionaware}. %
Differently from usual supervised training, where samples are drawn from a training set $\mathcal{T}$ that includes all classes $c \in \mathcal{C}$, we investigate a class-incremental scenario. 
In such a setup, the training is subdivided into $k = 0, \dots, K-1$ incremental steps. %
{At step $k$,} only the classes belonging to $\mathcal{C}_k \subset \mathcal{C}$ are provided with ground truth labels ($\mathbf{Y} \in \mathcal{C}_k^{H \times W}$). %
Similarly to previous works 
\cite{michieli2019,Michieli_2021_SDR,cermelli2020modeling,Camuffo_2023_CVPR}, we consider the \textit{overlapped} scenario, where at each step $k$ only the label distribution changes while the dataset remains the same. 
This is a forced choice for urban scenes where %
the sample separation needed in other incremental scenarios (\eg, \textit{sequential} or \textit{disjoint}) is unfeasible. Indeed, in road scenes, many classes are present within the same data sample (\eg \textit{road}, \textit{car}, or \textit{vegetation}), and 
{a strict label separation cannot be realistically implemented.}
Following the classes split proposed in \cite{Camuffo_2023_CVPR,klingner2020class}, we consider 5 incremental learning setups: \textbf{offline} (single step), \textbf{11-8} (2 steps), \textbf{6-5-8} (3 steps), \textbf{11-1}  (9 steps), \textbf{6-1} (14 steps).
Notation $k_1$-$k_2$ means that 
$k_1$ classes are shown at step $0$
and $k_2$ classes at each of the next steps until all the 19 classes are considered.

\section{Methodology}\label{sec:methodology}
In this section, we provide a detailed description of the model architecture as well as of the single modules that compose the overall system. 
The general model architecture is built departing from the multimodal architecture {PMF \cite{zhuang2021perceptionaware}.}
Such an approach is designed to learn the complementary features of color and LiDAR data (\ie, the appearance information from RGB images and the spatial information from LiDAR point clouds) via a two-stream architecture with residual-based fusion modules. The model is \rev{tuned} with a perception-aware loss that is able to \rev{capture} the difference between the \rev{outputs of the} two modalities \cite{technologies10040090}. 
The output of the network is a tuple of distinct semantic predictions that are used for optimization through multiple losses. %
However, {PMF's} structure is designed in such a way that it is \textit{not symmetric} and makes the LiDAR prediction the main result, as the residual connections are all oriented from the RGB branch towards the LiDAR one.

This represents a huge limitation \cut{of PMF} because LiDAR predictions are strongly affected by RGB data while \rev{the opposite is not true}. In addition, the model works only if both modalities are \cut{present}. %
Our approach is built on top of PMF and addresses such limitations to make the model \textit{fully-multimodal} and able to work well even in the absence of \cut{one sensor}.

\subsection{Multimodal Architecture}
As already pointed out we exploited a two-branch (one for LiDAR and one for color images) architecture based on PMF \cite{zhuang2021perceptionaware}.
We modified the architecture to make our model able to deal with unimodal scenarios on top of the multimodal \cut{one}. 
This choice is {motivated by many safety issues (\eg, the need for a seamless classification) and realistic conditions (\eg, differential slipping in the sensor synchronization) that may cause some missing acquisition for one of the sensors. \cut{Our scheme works even if one modality is not available.}

\textbf{Symmetric-Model structure.}
Figure \ref{fig:pipeline} represents the whole pipeline of our model. We moved from an asymmetric architecture (LiDAR as primary information and RGB data as side information) to a symmetric one (balanced across LiDAR and RGB information), developing a two-branch architecture that {jointly} outputs LiDAR and color predictions.
Our model takes as input the RGB image and the projected LiDAR point cloud. The projection is made in such a way as to make the point cloud aligned with its RGB counterpart.
In detail, LiDAR samples are projected in 2D space via perspective transformation {aligning their point of view to the one of the camera as in \cite{zhuang2021perceptionaware,cortinhal2020salsanext}}. These 2D features contain 5 values for each pixel position $(d,x,y,z,r)$, %
where $d = \sqrt{x^2+y^2+z^2}$ is the distance of a point from the camera, $(x,y,z)$ are the 3D point coordinates and $r$ is the {associated} reflectance.
A similar procedure can be employed on the segmentation labels, obtaining 2D remappings. We remark that, since the padding values (pixels where no LiDAR points fall into) are set to unlabeled and ignored during optimization, such 2D labels are also suitable to optimize the image branch. An example of a mapped point cloud is reported in the bottom half of Figure \ref{fig:inpaint}.

\textbf{Feature-Alignment.} 
To make the architecture symmetrical and aligned, we replace the self-attention-based fusion modules of PMF \cite{zhuang2021perceptionaware} with weighted sums of feature vectors, at 4 different feature levels:
\begin{equation} %
\mathbf{F}_{i} = r\mathbf{F}_{C,i} + (1-r)\mathbf{F}_{L,i}, \quad r \in [0,1]
\end{equation}
where $\mathbf{F}_{C,i}$ is the $i$-th feature generated by the color branch, $\mathbf{F}_{L,i}$ is $i$-th feature generated by the LiDAR branch, $\mathbf{F}_{i}$ is the combined feature, $r$ is a parameter that balances the two data sources and $i=0,1,2,3$ defines the depth level.
Note that $\mathbf{F} \in \mathbb{R}_0^+$ due to ReLU activations used in the network.

To enforce  feature alignment, we also define an additional loss function applied at each intermediate feature level:
\begin{equation} \small
    \mathcal{L}_{F} = \sum\limits_{i={0}}^{3} {||\mathbf{F}_{C,i} - \mathbf{F}_{L,i}||_2 + (1-\Theta(\mathbf{F}_{C,i},\mathbf{F}_{L,i}))}
\end{equation}
where $\Theta$ indicates the cosine similarity, \ie, the inner product between the normalized inputs. 
Note that we want to maximize the similarity, as such we minimize its negation. The rest of the supervised optimization is performed as in %
\cite{zhuang2021perceptionaware}.

\subsection{Continual Learning.}
Next, we introduce Continual Learning strategies {that preserve the functionality of the model} making it work also in challenging scenarios where some classes seen in the past are not present in the current training data. The following subsections describe in detail the problem setup and the relative modules.

\textbf{Knowledge Distillation (KD)}
Firstly, we apply knowledge distillation at the output level, to allow multimodal learning in different class-incremental continual learning settings. The multimodal {setup poses} the question of \textit{which supervision} should be used, and \textit{for which output}. 
{Indeed}, our model is designed to provide two predictions (one from the RGB branch and one from the LiDAR branch), and effectively work even in the case when the input is unimodal. Consequently, we must design Knowledge Distillation to cope with this \cut{setting}. %
  
We design four versions of the KD loss, according to the degrees of freedom of the system, exploring all the possible combinations.
We develop \rev{them} starting from a standard cross-entropy-based distillation loss\rev{:}
\begin{equation} \small
    \mathcal{L}_{KD}(\hat{\mathbf{Y}}_{k-1}, \hat{\mathbf{Y}}_{k}) = -\sum \hat{\mathbf{Y}}_{k-1}\log \hat{\mathbf{Y}}_{k}
\end{equation}
where $\hat{\mathbf{Y}}_{k-1}$  are the predicted class probabilities by the previous step ($k-1$) model and $\hat{\mathbf{Y}}_{k}$ are the predictions of current step ($k$) model.
The four losses are built on top of this general formulation. The difference between one another lies in the output and supervision provided.  We define them as follows:
{\small
\begin{align}
    \mathcal{L}_{KD,same} = & \mathcal{L}_{KD_{C, C}} + \mathcal{L}_{KD_{L, L}} \\
    \mathcal{L}_{KD,img} = & \mathcal{L}_{KD_{C, C}} + \mathcal{L}_{KD_{L, L}} + \mathcal{L}_{KD_{C, L}} \\
    \mathcal{L}_{KD,pcd} = & \mathcal{L}_{KD_{C, C}} + \mathcal{L}_{KD_{L, L}} + \mathcal{L}_{KD_{L, C}} \\
    \mathcal{L}_{KD,cross} = & \mathcal{L}_{KD_{C, C}} + \mathcal{L}_{KD_{L, L}} + \mathcal{L}_{KD_{C, L}} + \mathcal{L}_{KD_{L, C}}
\end{align}
}
where we omitted the arguments ($\mathbf{\hat{Y}}_{x,k-1}, \mathbf{\hat{Y}}_{y,k}$) for clarity and $\mathcal{L}_{KD_{x , y}}, x,y \in \{C,L\}$ is the KD loss, $x$ defines the supervision modality, and $y$ the supervised one. The compounded loss names are defined according to the modality used for supervision.
A general schematization of the losses can be found in Figure \ref{fig:pipeline}.
Effectively, with our knowledge distillation strategy, we can propagate the knowledge of old classes to the new model while bridging the gap between the two modalities. Notice that the first loss ($\mathcal{L}_{KD,same}$) just propagates the old knowledge inside each of the two modalities independently, while the second and third losses also from one modality to the other, while the last one performs a cross-supervision between them.
As it will be evident from the results in \rev{Sec.~\ref{sec:results}}, no single strategy is always the best for all classes. Nevertheless, $\mathcal{L}_{KD,same}$ and $\mathcal{L}_{KD,pcd}$ are the ones that {averagely obtain the best results} for LiDAR and color branches, respectively.

\textbf{Knowledge Inpainting.}
In addition to Knowledge Distillation, we exploit inpainting masks to produce reliable knowledge retention in a continual learning setting. For this purpose, we leverage our parallel-branch architecture to have access to the predictions of both modalities $\mathbf{\hat{Y}}_{C,k-1}$ and $\mathbf{\hat{Y}}_{L,k-1}$, and use both of them to estimate a more consistent multimodal prediction by averaging: $\mathbf{\hat{Y}}_{k-1} = \frac{1}{2}(\mathbf{\hat{Y}}_{C,k-1}+\mathbf{\hat{Y}}_{L,k-1})$.
After computing the probability average we compute the $\text{argmax}(\cdot)$ along the class dimension for each pixel location, effectively obtaining a tensor with the same shape as the labels $\mathbf{Y}$ {(referenced as pseudo-labels)}.
With these pseudo-labels, we inpaint the ground truth of step $k$ by overriding the labels of \textit{unknown} class {(\textit{i.e.}, the missing label information in the current step).} An example of this process is reported in Figure \ref{fig:inpaint}.

\begin{figure}[t]
    \centering 
    \begin{subfigure}{.3\textwidth}
    \rotatebox{90}{\hspace{0.2cm} \scriptsize Input}
    \includegraphics[width=\textwidth]{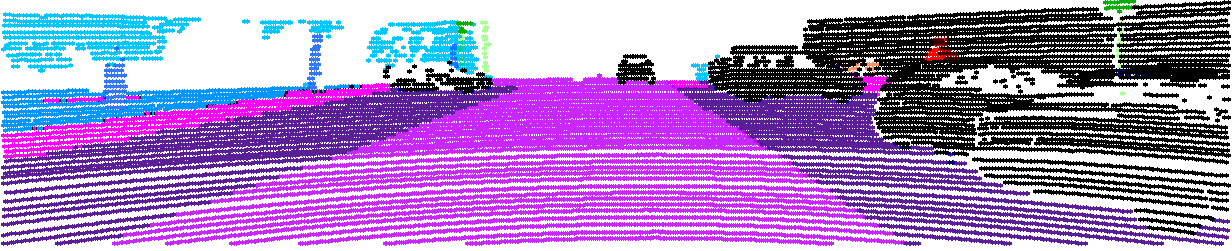}
    \end{subfigure}
    \begin{subfigure}{.3\textwidth}
    \rotatebox{90}{\hspace{0.1cm} \scriptsize Output}
    \includegraphics[width=\textwidth]{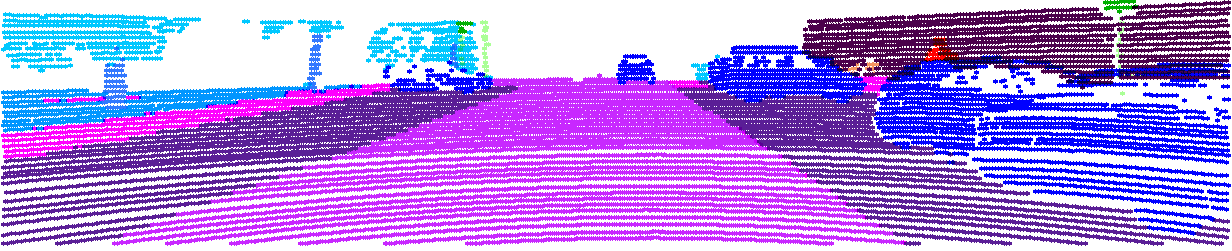}
    \end{subfigure} 
    \caption{\small Example of inpainting process, the \textit{unknown} class pixels (black) of the input image are filled in using the prediction of the network obtaining the inpainted output image. Points are expanded with a circular kernel for clarity, real size is one pixel. %
    }
    \label{fig:inpaint}
\end{figure}

\begin{table}[tb!]
    \setcounter{table}{0}
    \centering \footnotesize\renewcommand{\tabcolsep}{2pt}
    \begin{tabular}{cc|cc}
        \textbf{Input Modality} & \textbf{Branch} & \textbf{PMF} \cite{zhuang2021perceptionaware} & \textbf{Ours} \\
        \midrule
        \multirow{2}{*}{RGB+LiDAR} & RGB & 46.8 & \textbf{55.2} \\
                                 & LiDAR & \textbf{63.7} & 62.1 \\
        \hdashline
        \multirow{2}{*}{RGB} & RGB & \textbf{46.8} & 41.6 \\
                             & LiDAR & 16.1 & \textbf{57.2} \\
        \hdashline
        \multirow{2}{*}{LiDAR} & RGB & 1.4 & \textbf{48.0} \\
                               & LiDAR & 28.6 & \textbf{54.7} \\
        \hline
        \multirow{2}{*}{Average} & RGB & 31.6 & \textbf{48.2} \\
                               & LiDAR & 36.1 & \textbf{58.0} \\
    \end{tabular}
    \caption{\small Comparison in terms of mIoU (\%, $\uparrow$) between the original (asymmetric) PMF \cite{zhuang2021perceptionaware} formulation and our fail-safe symmetric approach. The last entry contains the average of the three cases at the top, providing a summarized result for easier comparison.}
    \label{tab:modal}
\end{table}

\begin{table*}[t]
    \setcounter{table}{1}
    \centering \footnotesize
    \setlength{\tabcolsep}{2.2pt}
    \begin{tabular}{ccc|ccccccccccccccccccc|c}
    \multicolumn{2}{c}{Method} & Modality & \rotatebox{90}{Road} & \rotatebox{90}{Parking} & \rotatebox{90}{S-walk} & \rotatebox{90}{Ot-Ground} &\rotatebox{90}{Vegetation} & \rotatebox{90}{Terrain} & \rotatebox{90}{Building} & \rotatebox{90}{Fence} & \rotatebox{90}{Trunk} & \rotatebox{90}{Pole} & \rotatebox{90}{T-Sign} & \rotatebox{90}{Bicycle} & \rotatebox{90}{M-cycle} & \rotatebox{90}{Truck} & \rotatebox{90}{Ot-Vehicle} & \rotatebox{90}{Person} & \rotatebox{90}{Bicyclist} & \rotatebox{90}{M-cyclist} & \rotatebox{90}{Car} & \rotatebox{90}{\textbf{mIoU}} \\
    \midrule 
    \multicolumn{2}{c}{\multirow{2}{*}{Offline}} & RGB & 95.7 & 29.8 & 78.5 & 0.0 & 86.0 & 72.2 & 85.0 & 55.9 & 63.3 & 62.5 & 39.2 & 19.9 & 45.1 & 32.3 & 55.3 & 65.9 & 68.4 & 0.0 & 93.2 & 55.2 \\
    & & LiDAR & 96.1 & 39.1 & 80.0 & 0.0 & 88.0 & 72.7 & 87.5 & 57.5 & 71.4 & 66.0 & 43.5 & 42.7 & 57.9 & 51.0 & 72.9 & 76.9 & 79.5 & 0.0 & 96.1 & 62.1 \\
    \hline
    \multirow{2}{*}{Ours} & \multirow{2}{*}{$\mathcal{L}_{KD,same}$} & RGB & \blue{94.9} & 37.3 & 75.0 & \blue{0.6} & 77.7 & 70.1 & 67.9 & 27.8 & \blue{41.7} & \blue{16.3} & \blue{31.7} & 1.1 & 0.3 & 1.7 & \blue{36.7} & 5.8 & 24.6 & \blue{0.0} & \blue{82.5} & 36.5\\
    & & LiDAR & \red{95.5} & 36.2 & 76.6 & \red{0.6} & \red{77.0} & 70.3 & 69.2 & 25.7 & \red{35.9} & \red{15.8} & 30.6 & 1.5 & 0.8 & 0.6 & 25.5 & 7.0 & 17.3 & \red{0.0} & \red{80.4} & \red{35.1}\\
    \hdashline
    \multirow{2}{*}{Ours} &  \multirow{2}{*}{$\mathcal{L}_{KD,img}$} & RGB & 94.5 & 35.5 & 75.1 & 0.5 & \blue{78.2} & 71.5 & 67.1 & 32.4 & 38.6 & 15.7 & 31.4 & 2.4 & 0.4 & 0.5 & 34.3 & 7.1 & 23.4 & \blue{0.0} & 81.8 & 36.3\\
    & & LiDAR & 95.3 & 35.7 & 76.2 & 0.3 & 76.5 & 71.4 & 67.5 & \red{32.1} & 30.6 & 13.9 & 29.4 & 1.9 & 0.5 & 0.1 & 24.7 & 7.7 & 15.9 & \red{0.0} & 76.0 & 34.5\\ 
    \hdashline
    \multirow{2}{*}{Ours} &  \multirow{2}{*}{$\mathcal{L}_{KD,pcd}$} & RGB & 94.7 & \blue{37.4} & \blue{75.6} & 0.5 & 77.4 & \blue{71.6} & \blue{69.7} & \blue{34.0} & 29.6 & 15.3 & 29.1 & \blue{2.5} & 0.5 & \blue{4.6} & 27.0 & \blue{10.5} & 20.2 & \blue{0.0} & 81.8 & \blue{36.9} \\
    & & LiDAR & 95.4 & \red{38.4} & \red{76.9} & 0.5 & 76.9 & \red{71.7} & \red{69.8} & 30.3 & 25.7 & 14.0 & 28.6 & 2.0 & 1.8 & \red{1.3} & 16.9 & \red{9.2} & 14.2 & \red{0.0} & 79.2 & 34.4\\
    \hdashline
    \multirow{2}{*}{Ours} & \multirow{2}{*}{$\mathcal{L}_{KD,cross}$} & RGB & 94.3 & 33.5 & 74.1 & \blue{0.6} & 73.9 & 70.2 & 61.6 & 29.6 & 26.7 & 9.3 & 31.5 & 2.1 & \blue{4.2} & 1.3 & 29.3 & 4.7 & \blue{25.4} & \blue{0.0} & 82.1 & 34.4\\
    & & LiDAR & 95.1 & 33.0 & 75.8 & 0.5 & 73.8 & 70.0 & 62.5 & 26.7 & 25.6 & 12.0 & \red{30.7} & \red{2.5} & \red{4.2} & 1.0 & \red{26.9} & 2.7 & 19.5 & \red{0.0} & 80.0 & 33.8\\
    \end{tabular} %
    \caption{\small Per-Class results mIoU (\%, $\uparrow$) on the last incremental step of \textbf{6-1} scenario. LiDAR best in \red{red} RGB best in \blue{blue}.}
    \label{tab:last}
\end{table*}

\begin{figure*}[!ht]
\centering
    \begin{minipage}[c]{\linewidth}
    \hspace*{-6em}
    \includegraphics[trim=6cm 22.58cm 0cm 0cm,clip,width=\linewidth]{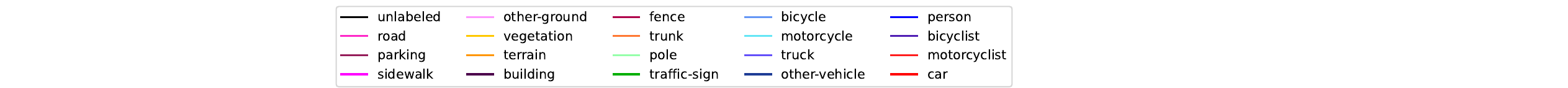}
    \end{minipage}\\
\begin{minipage}{.63\textwidth}
    \begin{tabular}[c]{cc}
  \rotatebox{90}{\hspace{-0.1cm} \scriptsize \textbf{IoU}}
     \begin{minipage}[c]{0.44\textwidth}
         \subcaption*{\scriptsize LiDAR \hspace{1.8cm} RGB}
      \includegraphics[trim=5.2cm 2cm 4.8cm 2.9cm,clip,width=\textwidth]{figs/plots/recall_6-1b_kd_cross.pdf}
      \subcaption{\footnotesize \textbf{6-1} $\mathcal{L}_{KD,cross}$}
      \label{fig:6-1cross}
    \end{minipage}&
   \begin{minipage}[c]{0.44\textwidth}
       \subcaption*{\scriptsize LiDAR \hspace{1.8cm} RGB}
      \includegraphics[trim=5.2cm 2cm 4.8cm 2.9cm,clip,width=\textwidth]{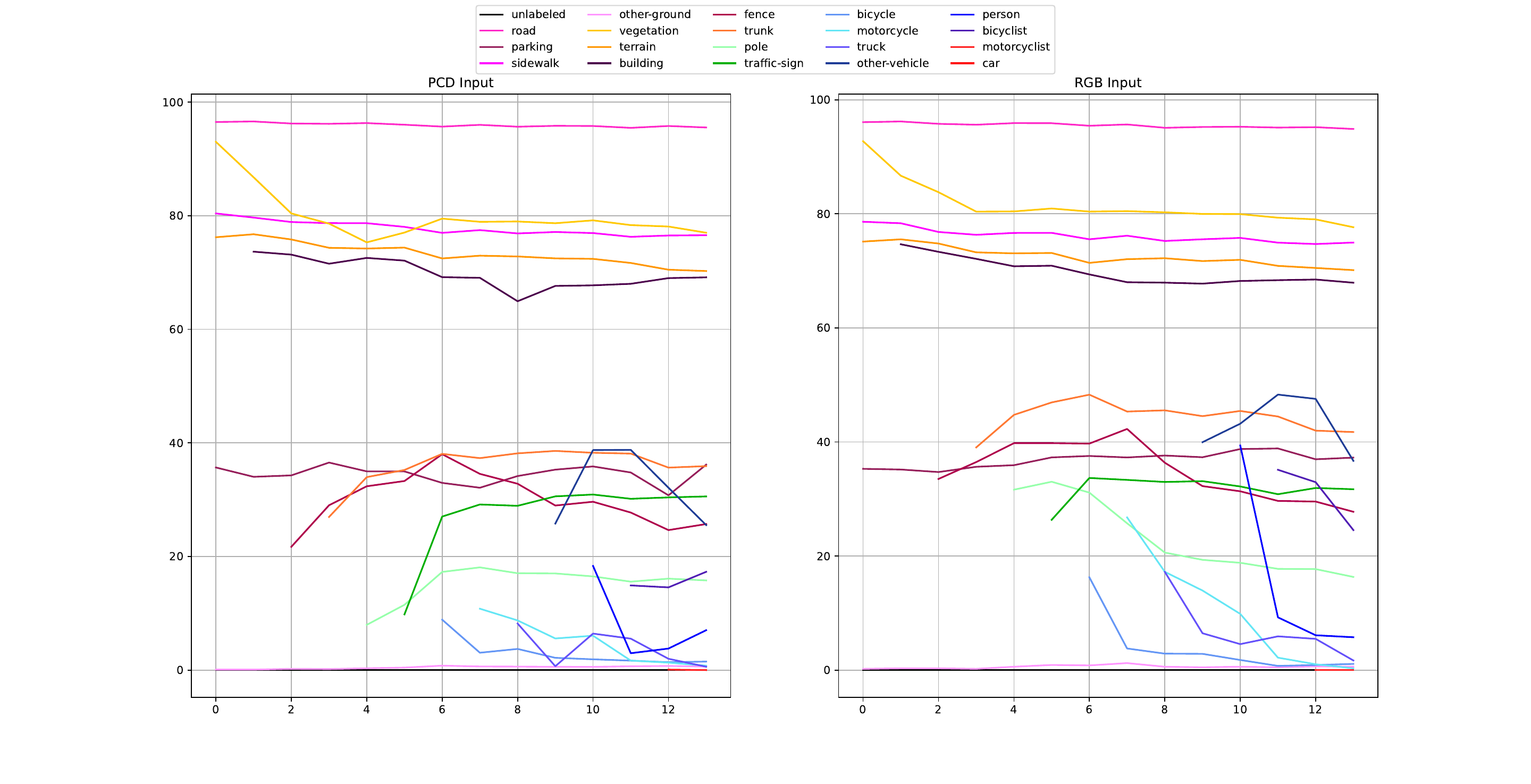}
      \subcaption{\footnotesize \textbf{6-1} $\mathcal{L}_{KD,same}$}
      \label{fig:6-1same}
    \end{minipage}\\\hfill 
    \rotatebox{90}{\hspace{-0.1cm} \scriptsize \textbf{IoU}}
        \begin{minipage}[c]{0.44\textwidth}
            \subcaption*{\scriptsize LiDAR \hspace{1.8cm} RGB}
      \includegraphics[trim=5.2cm 2cm 4.8cm 2.9cm,clip,width=\textwidth]{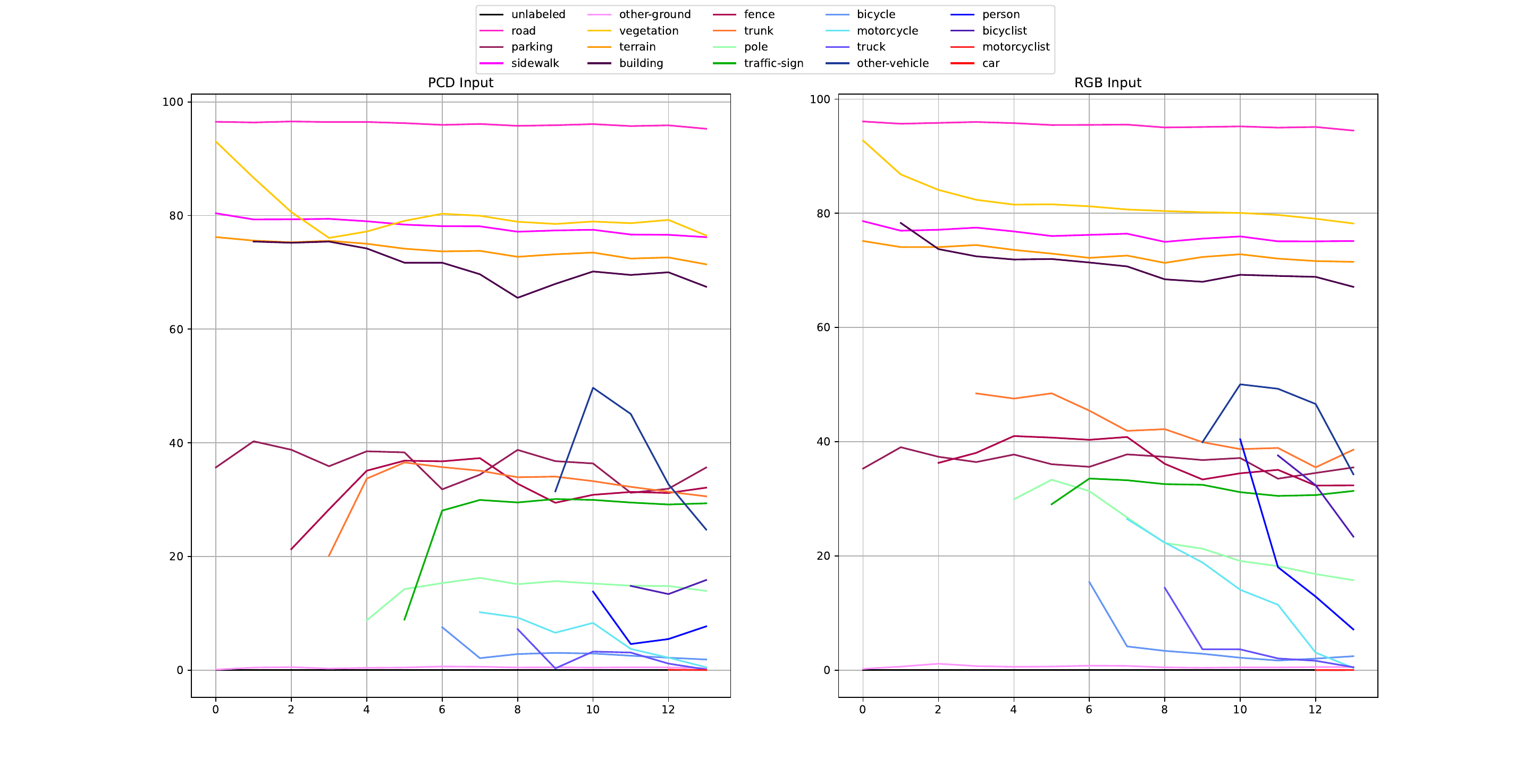}
       \subcaption{\footnotesize \textbf{6-1} $\mathcal{L}_{KD,img}$}
      \label{fig:6-1img}
   \end{minipage}&
       \begin{minipage}[c]{0.44\textwidth}
           \subcaption*{\scriptsize LiDAR \hspace{1.8cm} RGB}
      \includegraphics[trim=5.2cm 2cm 4.8cm 2.9cm,clip,width=\textwidth]{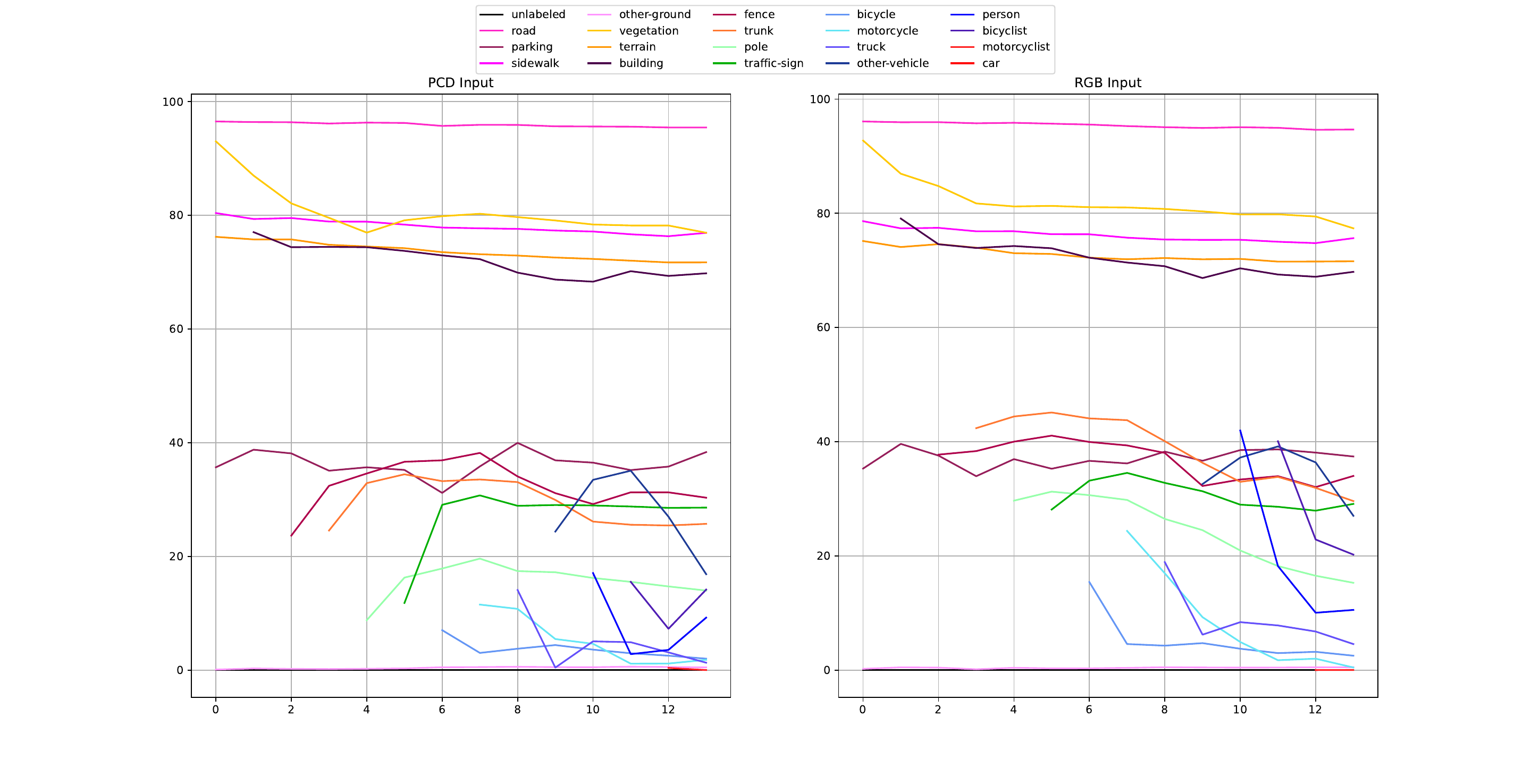}
       \subcaption{\footnotesize \textbf{6-1} $\mathcal{L}_{KD,pcd}$}
     \label{fig:6-1pcd}
    \end{minipage}\\
  \end{tabular}
\end{minipage}\hspace{1em}
\begin{minipage}{.34\textwidth}
    \centering \footnotesize
    \renewcommand{\tabcolsep}{2pt}
    \vspace{-5em}\begin{tabular}{ccc|cc} 
        \textbf{Task} & \textbf{Loss} & \textbf{Modality} & \textbf{Avg.} & \textbf{Per-Class} \\
        \midrule
        \multirow{8}{*}{\textbf{6-1}} & \multirow{2}{*}{$\mathcal{L}_{KD,same}$} & RGB & 2.04 & 2.04 \\
        & & LiDAR & 2.20 & 2.20 \\
        & \multirow{2}{*}{$\mathcal{L}_{KD,img}$} & RGB & 2.05 & 2.05 \\
        & & LiDAR & 2.24 & 2.24 \\
        & \multirow{2}{*}{$\mathcal{L}_{KD,pcd}$} & RGB & 2.09 & 2.09 \\
        & & LiDAR & 2.25 & 2.25 \\
        & \multirow{2}{*}{$\mathcal{L}_{KD,cross}$} & RGB & 2.20 & 2.20 \\
        & & LiDAR & 2.30 & 2.30 \\
        \hdashline
        \multirow{4}{*}{\textbf{11-1}} & \multirow{2}{*}{$\mathcal{L}_{KD,same}$} & RGB & 2.20 & 2.20 \\
        & & LiDAR & 2.72 & 2.72 \\
        & \multirow{2}{*}{$\mathcal{L}_{KD,cross}$} & RGB & 2.44 & 2.44 \\
        & & LiDAR & 2.84 & 2.84 \\
        \hdashline
        \multirow{2}{*}{\textbf{11-8}} & \multirow{2}{*}{$\mathcal{L}_{KD,same}$} & RGB & 11.16 & 1.45 \\
        & & LiDAR & 15.63 & 1.95 \\
        \hdashline
        \multirow{2}{*}{\textbf{6-5-8}} & \multirow{2}{*}{$\mathcal{L}_{KD,same}$} & RGB & 8.19 & 0.63 \\
        & & LiDAR & 10.38 & 0.80 \\
    \end{tabular}
    \setcounter{table}{2}
    \captionof{table}{\small Average mIoU drop ($\downarrow$) in the incremental steps. Per-class drop is computed by dividing the average drop by the number of incremental classes in the incremental steps.}
    \label{tab:drop}
\end{minipage}
\setcounter{figure}{3}
\caption{\small IoU (\%, $\uparrow$) results with different configurations of the Knowledge Distillation Loss $\mathcal{L}_{KD}$ for task \textbf{6-1}. \rev{The plot shows the IoU for each class at each incremental step ($x$-axis) starting from the step in which the class is firstly introduced. } %
} %
\label{fig:6-1b}
\end{figure*}

\subsection{Implementation Details}\label{sec:details}
The implementation of this work is based on PMF's \cite{zhuang2021perceptionaware} original codebase. Specifically, we derived the majority of the structure to keep the two architectures as close as possible and allow fairer comparisons.
We trained our architecture for 240k iterations on 4 NVIDIA RTX3090 GPUs with batch size 8 per GPU. For offline and continual step $0$ training, the learning rate was scheduled according to a cosine annealing with a linear warmup (2400 iterations), where the peak was set to $10^{-3}$. For the incremental steps, instead, the learning rate was scheduled with a linear decay from $10^{-3}$ to $5 \times 10^{-4}$.
Following PMF's strategy we employed two different optimizers, one for the color (SGD) and one for the LiDAR (Adam) branch. Both optimizers had weight decay enabled with rate $10^{-5}$. The SGD optimizer also had Nesterov momentum with rate $0.9$ enabled.
Note that the Adam optimizer was only used in \textbf{offline} and continual step $0$, while in the incremental steps, we employed SGD also for the LiDAR branch. We preserve the same data augmentation strategies of PMF.

\section{Experimental Results}\label{sec:results}
In this section, we report the quantitative and qualitative results attained by our approach for the SemanticKITTI \cite{behley2019semantickitti} benchmark. Such results refer to the multimodal/unimodal investigation, as well as to the continual learning scenarios. 
{The proposed symmetric multimodal setup is compared with the original (asymmetric) PMF strategy to evaluate the performance improvements. As for continual learning,} we compare our {approach} with the only other work - to the best of our knowledge - on continual learning for LiDAR data \cite{Camuffo_2023_CVPR}. We remark that the settings investigated in this work (especially the \textbf{6-1} task), are harder than those reported in \cite{Camuffo_2023_CVPR} as the number of {considered} incremental steps is larger. This is even harder, considering the unbalanced nature of the input data and the multimodal setting.

\textbf{Results on Multi/Uni-Modal} In this subsection, we report the comparison between the original, asymmetric, PMF architecture and our work. 
The architectures are evaluated in three settings: (1) multimodal, (2) RGB only, and (3) LiDAR only. Recall that, since we have two branches, each setting produces {a} LiDAR-branch accuracy {value} and {a} RGB-branch accuracy {value separately}.
The strength of our approach is evident from the results, reported in Table \ref{tab:modal}. 
In the multimodal setting, our (simpler) symmetric approach can improve the performance of the RGB prediction by $8.4\%$ with a small cost of $1.6\%$ in the LiDAR branch. 
{The gain is even stronger in the unimodal setup}:
thanks to the tightly-coupled feature extraction, {our approach} obtains reasonable accuracy even on the missing modality output. This is fundamental in a safety-critical application since a system cannot know ahead of time which input modality has failed, and ignore the relative output. 

Comparing our approach to PMF results, we noticed that when the LiDAR modality is missing the color branch of PMF can still provide a reasonable output - thanks to the asymmetric architecture, where color information is processed independently from the LiDAR one.
On the other hand, {whenever} the color information is missing, the architecture completely {fails}, since it is not able to {reasonably react} to the missing information {disabling the zeroed input. This implies that even the processing coming from the unaffected modality is corrupted.}
Missing or corrupted data on the color branch is common in real-world scenarios \cut{(\eg,  dense fog, heavy rain).}

An interesting and unexpected result of our architecture is that the LiDAR prediction is more accurate when the LiDAR information is missing and vice-versa for the color prediction. 
This may be related to the fact that we trained the architecture only on multimodal data. Consequently, the network focuses mostly on the cross-modality information to make a prediction, looking for clues missing in each of the input modalities.
We also remark that the low result on the RGB-RGB scenario is because the training regime was optimized for the multimodal setting; thus, it is reasonable that it may be less performing in other settings. Employing an early stopping approach, the RGB-RGB gives $47.1\%$ as a result, much closer to the $48.0\%$ attained by the LiDAR-LiDAR scenario.

\newcommand{\imwidth}{.29\linewidth}
\begin{figure}[t]
    \captionsetup{font=small}
    \begin{subfigure}{\linewidth}
        \centering
        \rotatebox{90}{\hspace{0.2cm} \footnotesize  RGB}
        \begin{subfigure}{\imwidth}
            \includegraphics[width=\textwidth]{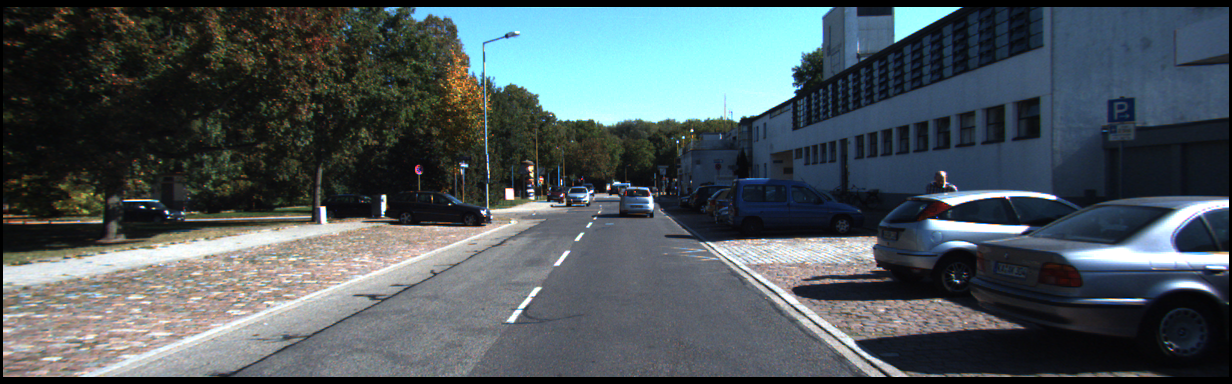}
        \end{subfigure}
        \begin{subfigure}{\imwidth}
            \includegraphics[width=\textwidth]{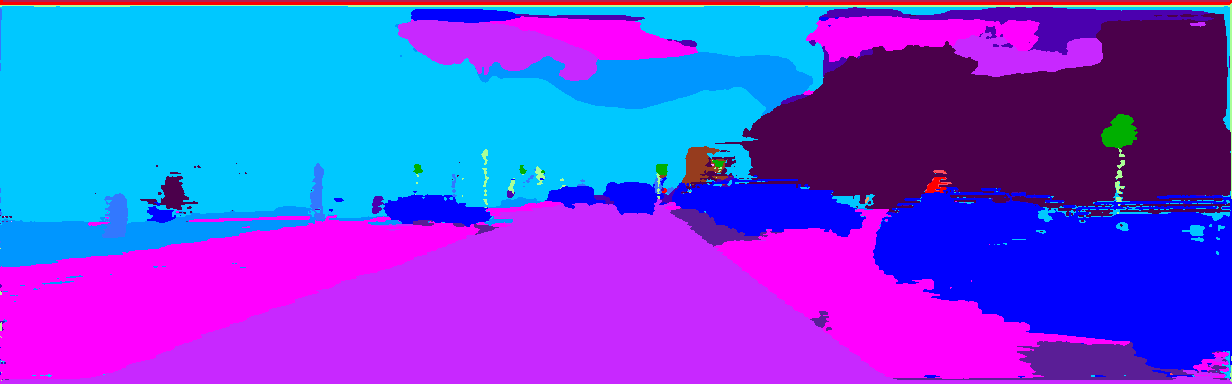}
        \end{subfigure}
        \begin{subfigure}{\imwidth}
            \includegraphics[width=\textwidth]{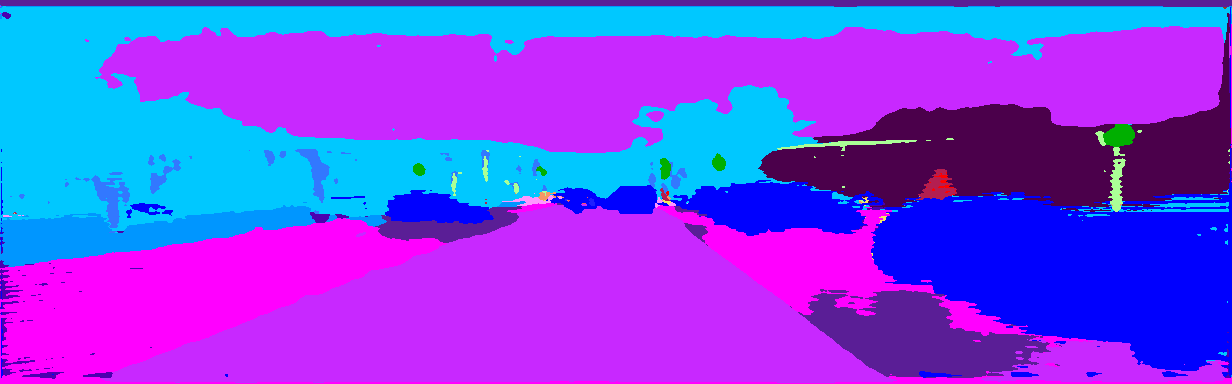}
        \end{subfigure}
    \end{subfigure}
    \begin{subfigure}{\linewidth}
        \centering
        \rotatebox{90}{\hspace{0.4cm} \footnotesize  LiDAR}
        \begin{subfigure}{\imwidth}
            \includegraphics[width=\textwidth]{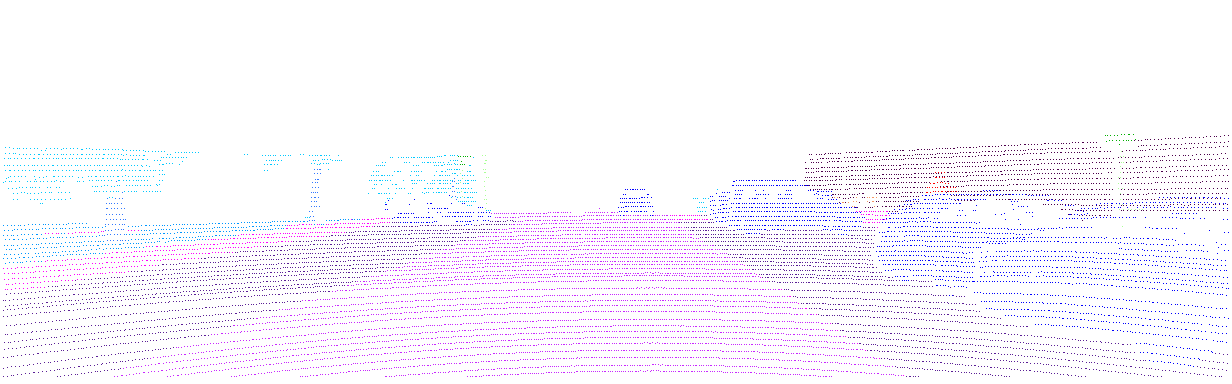}
            \caption*{Input/GT}
        \end{subfigure}
        \begin{subfigure}{\imwidth}
            \includegraphics[width=\textwidth]{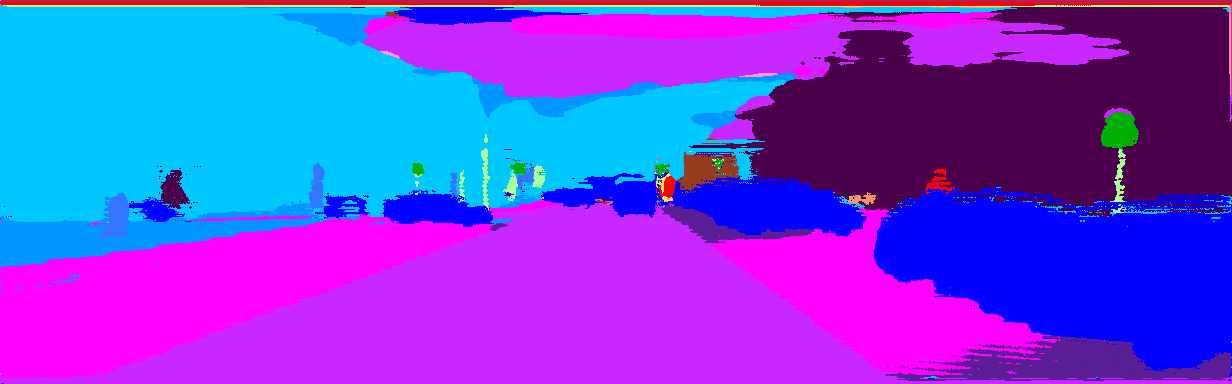}
            \caption*{Offline}
        \end{subfigure}
        \begin{subfigure}{\imwidth}
            \includegraphics[width=\textwidth]{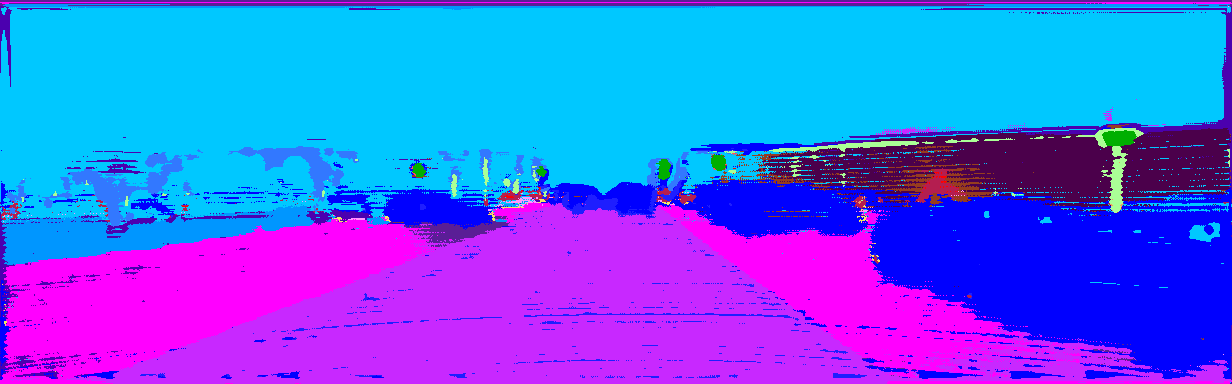}
            \caption*{$\mathcal{L}_{KD,same}$}
        \end{subfigure}
    \end{subfigure}
    \begin{subfigure}{\linewidth}
        \centering
        \rotatebox{90}{\hspace{0.2cm} \footnotesize  RGB}
        \begin{subfigure}{\imwidth}
            \includegraphics[width=\textwidth]{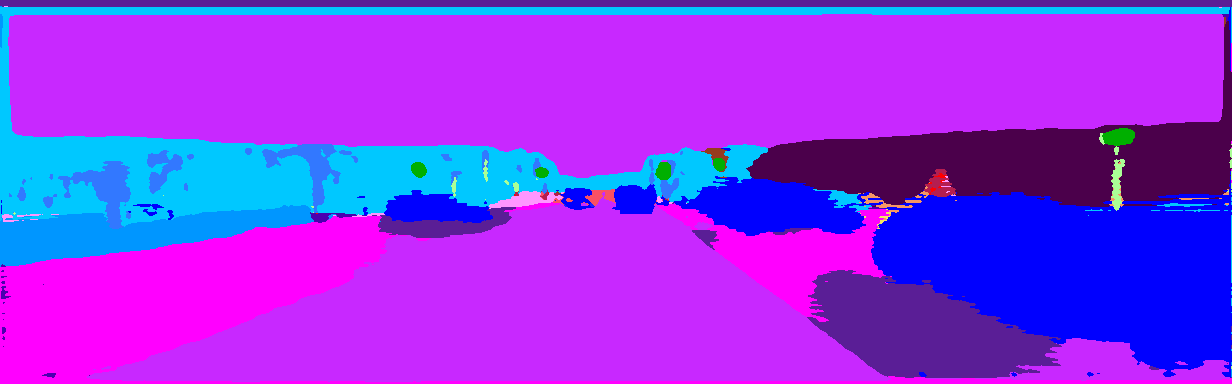}
        \end{subfigure}
        \begin{subfigure}{\imwidth}
            \includegraphics[width=\textwidth]{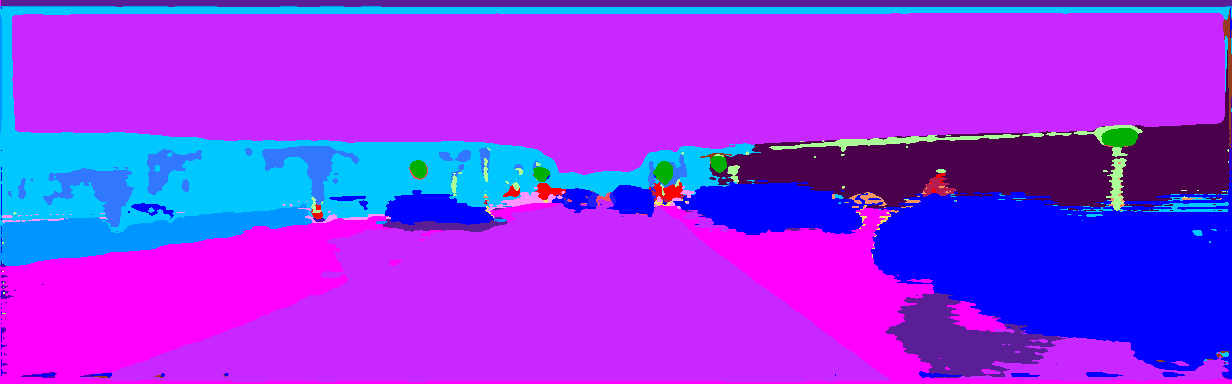}
        \end{subfigure}
        \begin{subfigure}{\imwidth}
            \includegraphics[width=\textwidth]{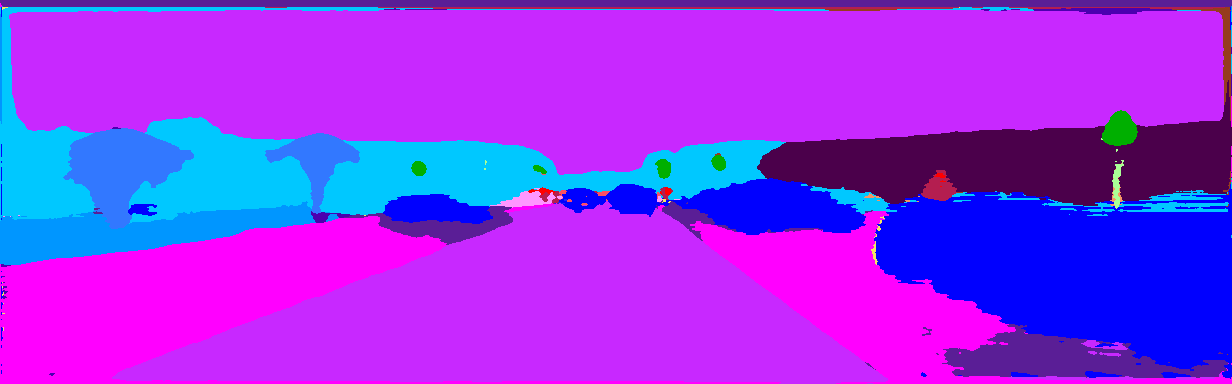}
        \end{subfigure}
    \end{subfigure}
    \begin{subfigure}{\linewidth}
        \centering
        \rotatebox{90}{\hspace{0.4cm} \footnotesize LiDAR}
        \begin{subfigure}{\imwidth}
            \includegraphics[width=\textwidth]{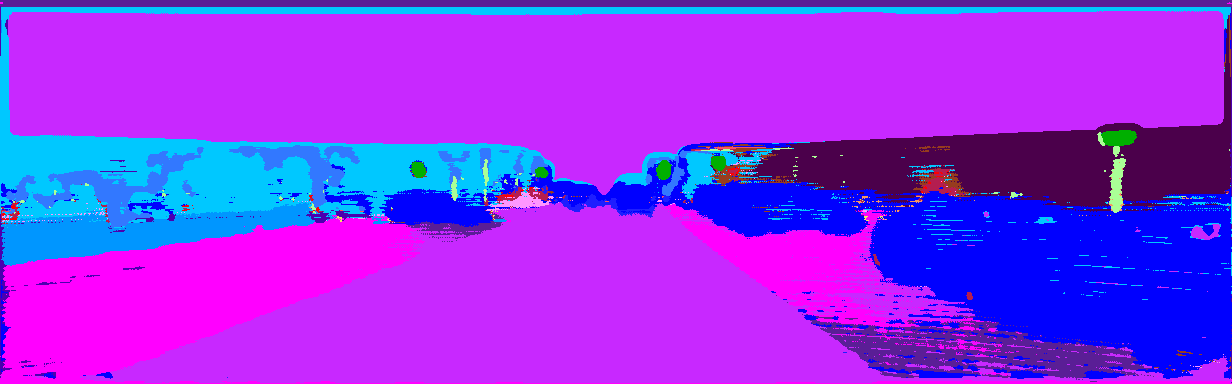}
            \caption*{$\mathcal{L}_{KD,img}$}
        \end{subfigure}
        \begin{subfigure}{\imwidth}
            \includegraphics[width=\textwidth]{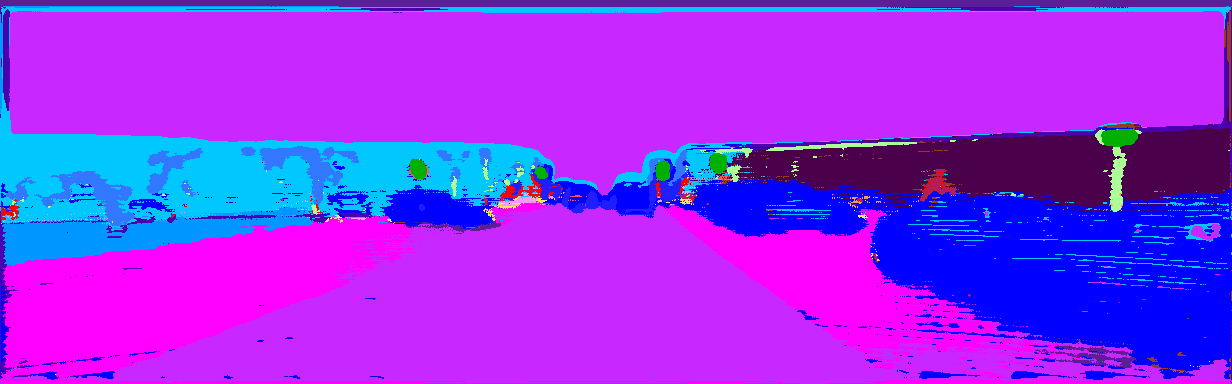}
            \caption*{$\mathcal{L}_{KD,pcd}$}
        \end{subfigure}
        \begin{subfigure}{\imwidth}
            \includegraphics[width=\textwidth]{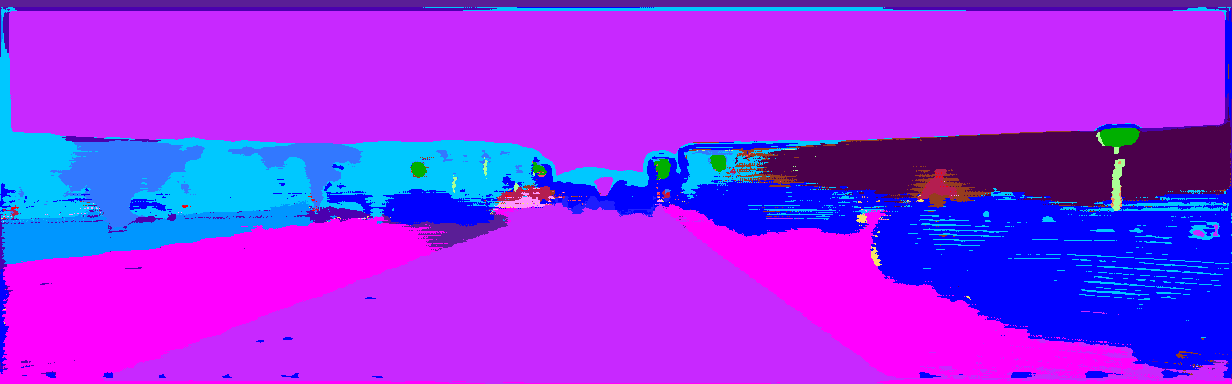}
            \caption*{$\mathcal{L}_{KD,cross}$}
        \end{subfigure}
    \end{subfigure}
    \caption{\small Qualitative results. The top rows contain the image branch predictions, bottom rows contain the LiDAR branch predictions.}
    \label{fig:quali}
\end{figure}
\textbf{Results on Continual Learning}
In this subsection, we report the quantitative results and ablation studies {related to the}  continual learning scenario.
Figure \ref{fig:6-1b} reports in summarized form the per-class per-step IoU for the \textbf{6-1} {setup} with the four different formulations of the KD loss.
Note that each figure shows two paired plots, one for LiDAR and one for RGB IoU outputs. 
From the figures, it is clear that in all configurations our feature-alignment strategy results in a very tight coupling between LiDAR and RGB performance. Moreover, our continual learning approaches can preserve in a significant manner the knowledge of old classes - especially those learned in the very first steps as they are (generally) the easiest, 
\rev{while the performance on some of the latter ones (\textit{e.g.} \textit{M-cycle} and \textit{Bicycle}) is lower, mostly due to their small size and limited number of samples.} Remarkably, our approach maintains \rev{in general stable performances}  both with $\mathcal{L}_{KD,cross}$ and $\mathcal{L}_{KD,same}$.\\
Table \ref{tab:last} reports some extended results for the last incremental step of scenario \textbf{6-1} in comparison with the \textbf{offline} (non-incremental) training.
Analyzing the table it is evident that there is no clear best formulation for the knowledge distillation loss, as different information-sharing strategies lead to {different} improvements {on the diverse classes}.
Each of the 4 cases can achieve the best performance on one of the different classes, both in the image and the LiDAR branches.
Overall, the highest LiDAR performance is attained by the $\mathcal{L}_{KD,same}$ loss, while the best RGB mIoU is found in the $\mathcal{L}_{KD,pcd}$ case. This result is in line with those reported in Table \ref{tab:modal} where the result is strongly affected by the cross-modal information employed by our architecture for the final prediction.

Table %
\ref{tab:drop} reports the average mIoU decrease per incremental step in all scenarios, confirming previous findings: the RGB prediction is more stable than the LiDAR one and this is reflected in lower performance degradation. 
Comparing task \textbf{11-1} and task \textbf{11-8}, we can notice that significantly higher performance can be obtained while showing several classes within the same incremental step: in the former task, the per average class drop (both for RGB and LiDAR) is 2.46\%,  in the latter is \rev{1.7\%}. This corresponds to an improvement of 44.71\%.

Figure \ref{fig:quali} reports qualitative results comparing various formulations of knowledge distillation strategy and offline learning. In general, we can notice that the LiDAR predictions are less smooth than the color ones, an artifact deriving from the sparse nature of the inputs and the multi-scale processing. Secondly, in all continual learning predictions, the cutoff in LiDAR projections - between points where LiDAR points exist and where they do not - is more evident than in the offline case. We can assume this spawns from the reduced complexity of the task that needs to be solved by the architecture. Indeed, in continual learning, much of the network capacity is used to maintain old knowledge, while in one-shot training the model focuses more on recognizing shapes and sharing information between the branches.
From the figures, it is also clear how the color information provides helpful clues to the LiDAR branches. For example, compare the person on the right side of the image in the \textit{KD Same LiDAR} and \textit{KD Img LiDAR}: in the former, the prediction is very noisy and the points bleed around the original silhouette, while in the latter, the definition is significantly higher and there is no bleeding.
Another interesting result can be noticed by looking at the bottom right of the KD,Img images, in those cases the network can correctly recognize the \textit{other-ground} class (purple), which is challenging to detect even in the \textbf{offline} scenario.

\vfill
\section{Conclusions}\label{sec:conclusions}
In this work %
\cut{ we propose} a LiDAR and RGB multimodal architecture that can work even in the presence of a single modality.
This is achieved by changing the asymmetrical model structure into a symmetrical one, and by introducing a tightly-coupled feature representation between the RGB and LiDAR branches, enforced by additional loss functions. 

Results show that our method reaches state-of-the-art results with an overall average gain of $16.6\%$ mIoU on the RGB predictions and $21.9\%$ mIoU on the LiDAR predictions with respect to PMF \cite{zhuang2021perceptionaware}, outperforming competitors, especially in the case of an unimodal setup.
Furthermore, we devise class-incremental continual learning on top of our model: we employ knowledge distillation and inpainting strategies to avoid \textit{catastrophic forgetting} %
\cut{applying} supervision on the same modality, on both or from one  to the other.
Results prove  our method  effectiveness also in a multimodal continual learning setup.

\newpage
\bibliographystyle{IEEEbib}
\bibliography{biblio}

\end{document}